\documentclass[letterpaper, 10 pt, conference]{ieeeconf}
\IEEEoverridecommandlockouts
\usepackage{epsfig} 
\usepackage{amsmath}
\usepackage{multirow}
\usepackage{tabularx}
\usepackage{color}
\usepackage{algorithm}
\usepackage{algorithmicx}
\usepackage{algpseudocode}
\usepackage{bm}
\usepackage{gensymb}
\usepackage{amsfonts}
\usepackage{graphics}
\usepackage{float}
\usepackage{url}
\usepackage{siunitx}
\usepackage{mathrsfs}
\usepackage{soul}
\usepackage{flushend}
\usepackage{caption}
\usepackage{subcaption}
\usepackage{booktabs}
\usepackage[dvipsnames]{xcolor}
\usepackage{hyperref}
\hypersetup{backref=true,
pagebackref=true,
hyperindex=true,
colorlinks=true,
breaklinks=true,
urlcolor= black,
linkcolor= blue,
bookmarks=true,
                    bookmarksopen=false,
                    citecolor=black,
                    linkcolor=black,
                    filecolor=black,
                    citecolor=blue,
                    linkbordercolor=blue
}

\algtext*{EndWhile}
\algtext*{EndIf}
\algtext*{EndFor}

\soulregister{\cite}7
\soulregister{\citep}7
\soulregister{\citet}7
\soulregister{\ref}7
\soulregister{\pageref}7
\soulregister{\uppercase\expandafter}7
\sethlcolor{yellow}

\newcommand{\deletefig}[1]{{\bgroup\markoverwith{\textcolor{red}{\rule[2.5ex]{2pt}{2.0pt}}}\ULon{#1}}}

\newcommand{\delete}[1]{}

\usepackage[misc]{ifsym}

\title{ \LARGE \bf
Learning Humanoid Locomotion with Perceptive Internal Model
}

\author{Junfeng Long$^*$, Junli Ren$^*$, Moji Shi$^*$, Zirui Wang, Tao Huang, Ping Luo, Jiangmiao Pang
\thanks{All authors are with Shanghai AI Laboratory. * denotes equal contribution.
Junli Ren and Ping Luo are with the University of Hong Kong. 
Zirui Wang is with Zhejiang University.
Tao Huang is with Shanghai Jiao Tong University.
Corresponding author: Jiangmiao Pang (pangjiangmiao@gmail.com). Project Page: \url{https://junfeng-long.github.io/PIM/}.
}
}

\begin{document}
\maketitle
\thispagestyle{empty}
\pagestyle{empty}

\begin{abstract}
In contrast to quadruped robots that can navigate diverse terrains using a ``blind" policy, humanoid robots require accurate perception for stable locomotion due to their high degrees of freedom and inherently unstable morphology. However, incorporating perceptual signals often introduces additional disturbances to the system, potentially reducing its robustness, generalizability, and efficiency. This paper presents the Perceptive Internal Model (PIM), which relies on onboard, continuously updated elevation maps centered around the robot to perceive its surroundings. We train the policy using ground-truth obstacle heights surrounding the robot in simulation, optimizing it based on the Hybrid Internal Model (HIM), and perform inference with heights sampled from the constructed elevation map. Unlike previous methods that directly encode depth maps or raw point clouds, our approach allows the robot to perceive the terrain beneath its feet clearly and is less affected by camera movement or noise. Furthermore, since depth map rendering is not required in simulation, our method introduces minimal additional computational costs and can train the policy in 3 hours on an RTX 4090 GPU. We verify the effectiveness of our method across various humanoid robots, various indoor and outdoor terrains, stairs, and various sensor configurations. Our method can enable a humanoid robot to continuously climb stairs and has the potential to serve as a foundational algorithm for the development of future humanoid control methods.

\end{abstract}

\begin{figure*}[!h]
\centering{\includegraphics[width=\textwidth]{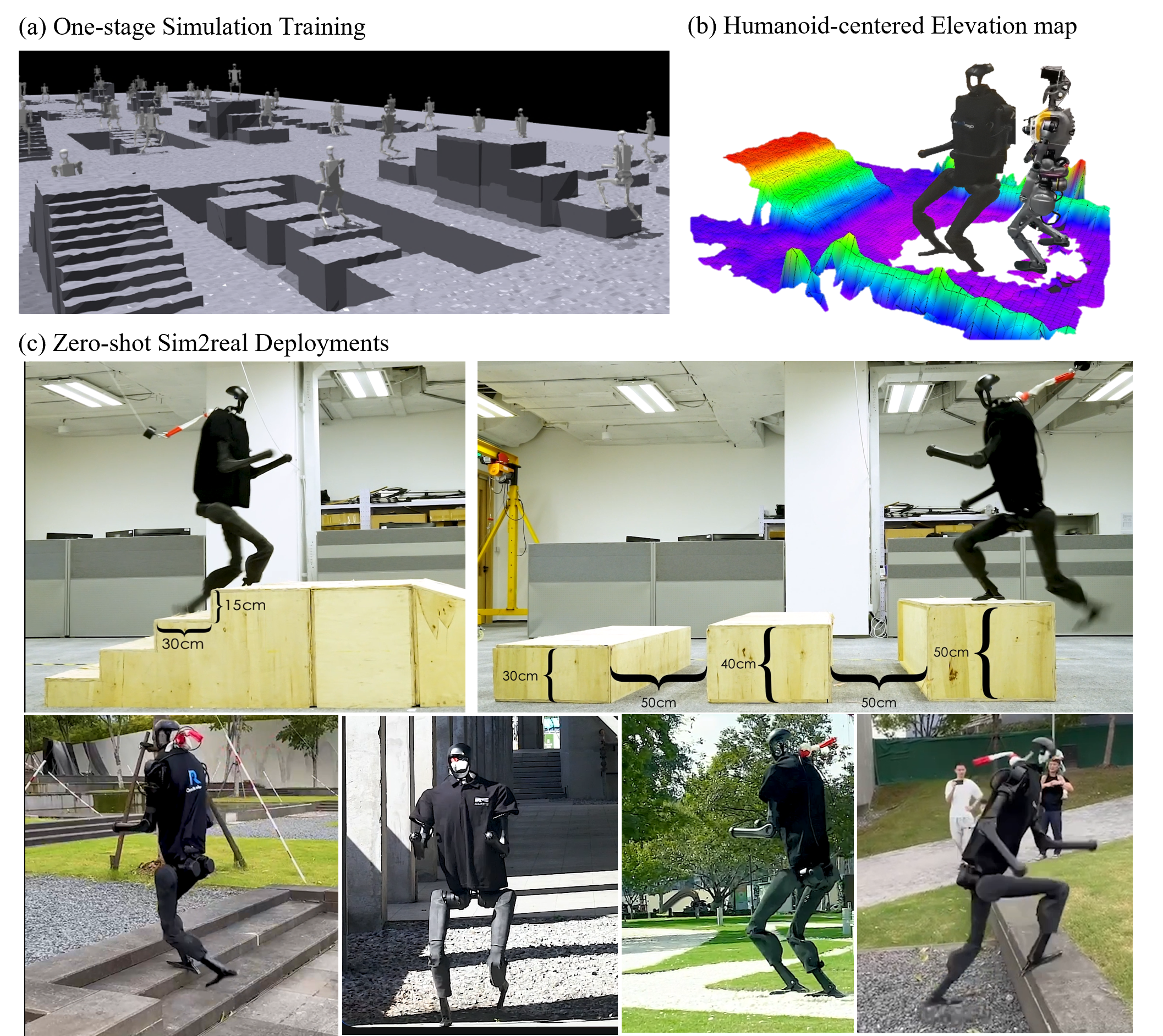}}
\caption{We propose a perceptive humanoid locomotion policy capable of mastering various challenging terrains. This policy requires only a few hours of single-stage training and can be deployed on hardware in a zero-shot manner. By integrating a LiDAR-based elevation map, we achieve accurate perception without the need for historical vision embedding. }
\label{Intro}
\end{figure*}
\section{Introduction} \label{Section : Introduction}

The control methods for legged robots have advanced rapidly in recent years ~\cite{hwangbo2019learning,lee2020learning}, driven by robot learning techniques such as reinforcement learning and simulation-to-real transfer. Among these methods, ``blind" policies, which rely solely on proprioceptive information as observations, have proven effective in enabling quadruped robots to achieve robust and agile locomotion across various terrains~\cite{long2024hybrid, long2024learning, margolis2023walk}.

Due to the higher degrees of freedom and inherently unstable morphology of humanoid robots, it is significantly more difficult to achieve stable locomotion across diverse terrains using a single policy, making the integration of perceptual information essential.
However, considering that incorporating perceptual signals may introduce additional disturbances and costs to the system, potentially compromising its robustness, generalizability, and efficiency, most recent efforts in humanoid locomotion have followed the ``blind" policy ~\cite{gu2024advancing,radosavovic2024humanoid,liao2024berkeley} or a two-phase training paradigm ~\cite{cuiadapting,zhuang2024humanoid}. 
They may rely on prior motion trajectories~\cite{cheng2024expressive, he2024learning, dugar2024learning} or multi-stage, computationally expensive training processes~\cite{cuiadapting} to execute human-like locomotion actions that can be deployed on real-world robots.
While these approaches successfully enable humanoid robots to walk on flat surfaces and mildly uneven terrains such as low stairs, they often end up falling and can not handle more challenging scenarios. 
Humanoid Parkour Learning~\cite{zhuang2024humanoid} exploits perceptive information but is not able to overcome terrains requiring fine-grained footholds such as continuous stairs.

This paper presents the Perceptive Internal Model (PIM), which relies on onboard, continuously updated elevation maps centered around the robot to perceive its surroundings.
It is built upon the Hybrid Internal Model (HIM)~\cite{long2024hybrid}, which uses batch-level contrastive learning to optimize the simulated robot's response, incorporating both explicit velocity and implicit latent, to the robot's successor state. 
To incorporate perceptive information, PIM directly uses the ground-truth obstacle height maps surrounding the robot as additional observations to train the policy in simulation. During inference, we construct the elevation map using a LiDAR or RGB-D camera and sample points from it to align with the policy's observations.

In contrast to previous methods that directly encode depth maps or raw point clouds, PIM advances in:
(a) It benefits from HIM, which supports batch-level learning, single-stage training, and high training efficiency.
(b) The elevation map accounts for the robot's odometry and continuously maintains a larger local map, allowing the robot to clearly perceive the terrain beneath its feet. Compared to raw sensory data, it is more robust to sensor movement and noise.
(c) Rendering depth images in simulation incurs additional costs in both computational memory and efficiency, whereas directly querying terrain heights does not. Furthermore, mitigating the domain gap between simulated depth maps and real camera data is challenging. As a result, our method introduces minimal additional computational costs and completes policy training within 3 hours on an RTX 4090 GPU.

We verify the effectiveness of our proposed method across various humanoid robots, including Unitree H1 and Fourier GR-1, on a range of challenging indoor and outdoor terrains, such as stairs, gaps, and high platforms, and with different sensor configurations, including Mid-360 LiDAR and RealSense D435+T265 cameras.
Our method enables humanoid robots to navigate a variety of difficult terrains with a natural gait and a high success rate. It achieves a success rate of over 90\% when continuously climbing stairs.
We hope that PIM can serve as a foundational algorithm for the development of future humanoid control algorithms.

\section{Related Work}  \label{sec:related-works}

\subsection{Learning Humanoid Locomotion}
The rise of the humanoid robot industry has motivated extensive research of humanoid locomotion in the past year~\cite{cheng2024expressive,cuiadapting,dugar2024learning,fu2024humanplus,gu2024advancing,gu2024humanoid,hansen2024hierarchical,he2024learning,jiangharmon,liao2024berkeley,liokami,radosavovic2024humanoid,van2024revisiting,zhang2024wococo,zhaobi,zhuang2024humanoid}, a majority of these researches follows the "training in simulation \& Sim2Real transfer" procedure, akin to the methods employed in quadruped research. Indeed, when it comes to humanoid robots, there is an emphasis on performing human-like movements~\cite{he2024learning,he2024omnih2o}. To achieve this,~\cite{cheng2024expressive,radosavovic2024humanoid,dugar2024learning} utilize prior motion trajectories to learn natural whole-body motion while walking on diverse flat grounds. Despite the expressive motions demonstrated in these studies, subsequent research has indicated that learning from scratch can also yield coordinated whole-body movement~\cite{zhang2024wococo,zhuang2024humanoid}, especially when the task is focused on traversing challenging terrains.

On the other hand, learning a robust legged locomotion policy requires the prediction of accurate robot states from accessible observations. Given that humanoid robots typically have higher dimensions of action and observation space, recent works have proposed different training diagrams to address this issue:~\cite{liao2024berkeley} incorporates linear velocity into observations from a separate state estimator,~\cite{zhuang2024humanoid, cuiadapting} encompasses multi-stage training process to narrow the sim2real gap,~\cite{gu2024advancing} predicts the states from observation as part of a denoising process. 

Unlike the previous works, we introduce a novel locomotion policy that accomplishes state prediction utilizing perceptive information, thereby eliminating the necessity for additional training stages or explicit decoding of the predicted states.

\subsection{Perceptive Legged Locomotion}
Previous research on quadrupeds has formulated several successful systems that introduce exteroceptive sensors into the locomotion policy~\cite{cheng2024extreme,agarwal2023legged,hoeller2024anymal,miki2022learning}. These policies, which either receive heightmaps or depth images, have demonstrated that integrating a more comprehensive vision system and vision observations into the policy results in better performance on challenging terrains~\cite{miki2022elevation}. In the context of humanoid robots, while the aforementioned blind policies have demonstrated robust performance on various surfaces ~\cite{radosavovic2024humanoid}, against disturbances~\cite{cheng2024expressive,he2024learning}, and on simple stairs~\cite{gu2024advancing}, a comprehensive vision-aided policy is necessary to achieve perceptive locomotion. A significant challenge in perceptive locomotion is reducing the perception gap during sim2real transfer. Existing research~\cite{zhuang2024humanoid} employs an extra phase of training to simulate the extensive hardware noise. However, this approach compromises perception accuracy and prevents the policy from accomplishing tasks that demand more precise perception, such as stair climbing.

Compared to existing methods, we propose a more comprehensive perceptive locomotion system that achieves more precise hardware perception by incorporating a LiDAR-based elevation map. The integration of a robot-centered elevation map eliminates the need for introducing vision into the historical observation sequence and achieves a more comprehensive state prediction. Extensive experiments on terrains involving stairs, gaps, and platforms underscore the superiority of the proposed method compared to existing studies.
\section{Methodology} \label{sec:methodology}

\begin{figure*}[h]
\centering{\includegraphics[width=\textwidth]{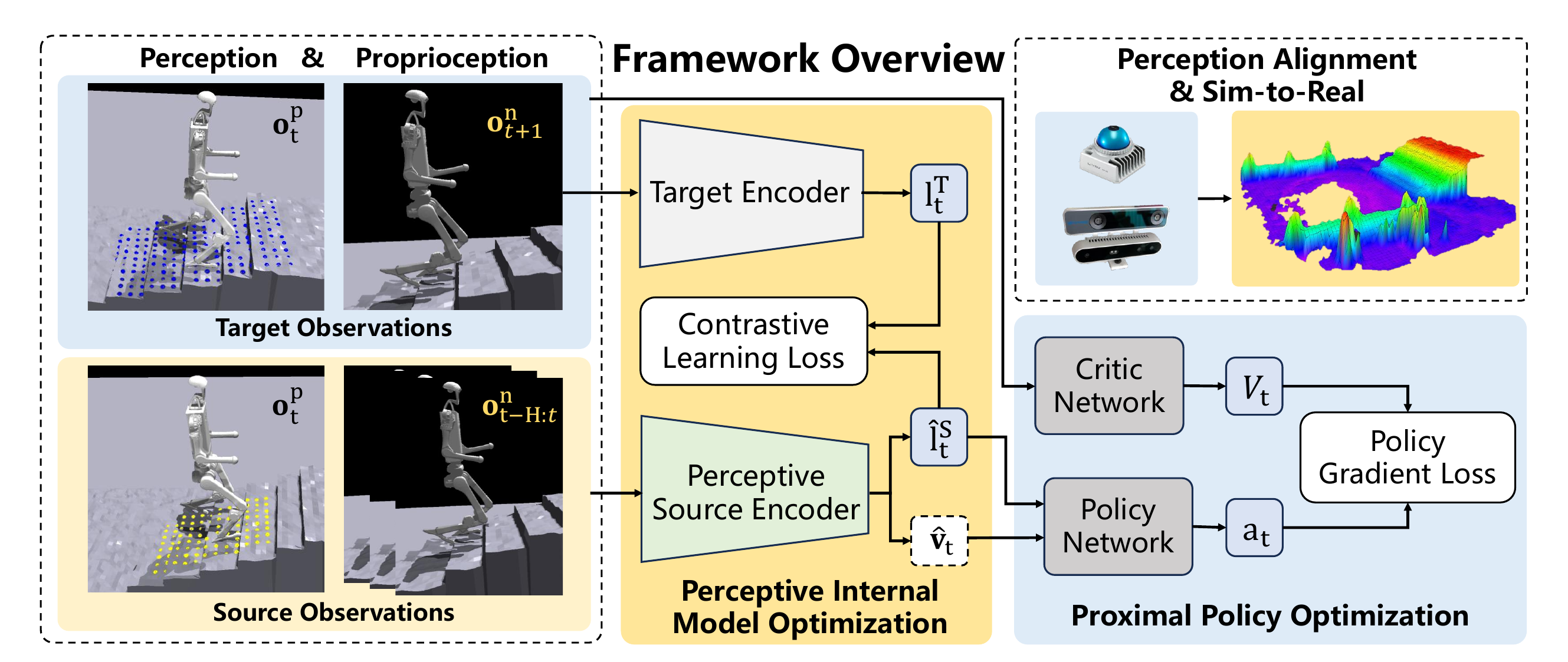}}
\caption{Overview of our framework. Within PIM, we integrate perceptive information into the state predictor to achieve more comprehensive and accurate state prediction. A LiDAR-based elevation map serves as the perception model, enabling more precise perception alignment between simulation and real-world environments.}
\label{method}
\end{figure*}

\subsection{Framework Overview}
Compared to quadruped robots, humanoid robots have significantly lower stability due to their morphology.
The perceptive information is necessary for the control system. 
It needs to determine the movements of its actuators for a desired velocity given its proprioceptive information and perceptive information. We model the humanoid locomotion task as a sequential decision problem. The entire framework is depicted in Fig. \ref{method}, we construct a perceptive internal model that utilizes both proprioceptive information and perceptive information for state estimation and optimize the policy with PPO~\cite{schulman2017proximal}. 

\subsection{Observations} 
The policy observations $\mathbf{o}_{t}$ consist of commands, proprioceptive information, perceptive information, and action of last timestep $\mathbf{a}_{t-1}$.
The commands are the desired velocity $\mathbf{c}_t = [v_x^c, v_y^c, \omega_{\text{yaw}}^c]$ indicates the linear velocity in longitudinal and lateral directions, and the angular velocity in the horizontal direction, respectively. The proprioceptive information includes its joint position $\mathbf{\theta}_t$, joint velocity $\dot{\mathbf{\theta}}_t$, base angular velocity $\mathbf{\omega}_t$ and gravity direction in robot frame $\mathbf{g}_t$. The perceptive information $\mathbf{p}_{t}$ is an elevation map around the robot. We divide these observations into two parts, non-perceptive part $\mathbf{o}^{n}_{t}$ and perceptive part $\mathbf{o}^{p}_{t}$, i.e. $\mathbf{o}^{n}_{t} = [\mathbf{c}_t, \mathbf{\omega}_t, \mathbf{g}_t, \mathbf{\theta}_t, \dot{\mathbf{\theta}}_t, \mathbf{a}_{t-1}]$ and $\mathbf{o}^{p}_{t} = [\mathbf{p}_t]$. 
Unlike the policy, the critic is allowed to access privileged information such as the linear velocity $v_{t}$ of the robot at the training stage and provide a more accurate estimation of state values.

\subsection{Preliminary: Hybrid Internal Model}
HIM (Hybrid Internal Model)~\cite{long2024hybrid} provides accurate state estimation and sim2real ability. Basically, HIM estimates the robot's linear velocity in the next timestep and proprioceptive information in a latent space given its proprioceptive observation history. The linear velocity estimation is trained with simple regression to the ground truth linear velocity obtained from the simulator. The next step proprioceptive information prediction is trained through contrastive learning. HIM considers a pair of $\mathbf{o}^{n}_{t+1}$ and $\mathbf{o}^{n}_{t-H:t}$ from the same trajectory as a positive pair and others as negative pairs.
The pairs are optimized by swapping assignments tasks similar to SwAV\cite{caron2020unsupervised}.  

\subsection{Perceptive Internal Model}

\paragraph{Elevation Maps}
\begin{figure}
    \centering
    \includegraphics[width=0.8\linewidth]{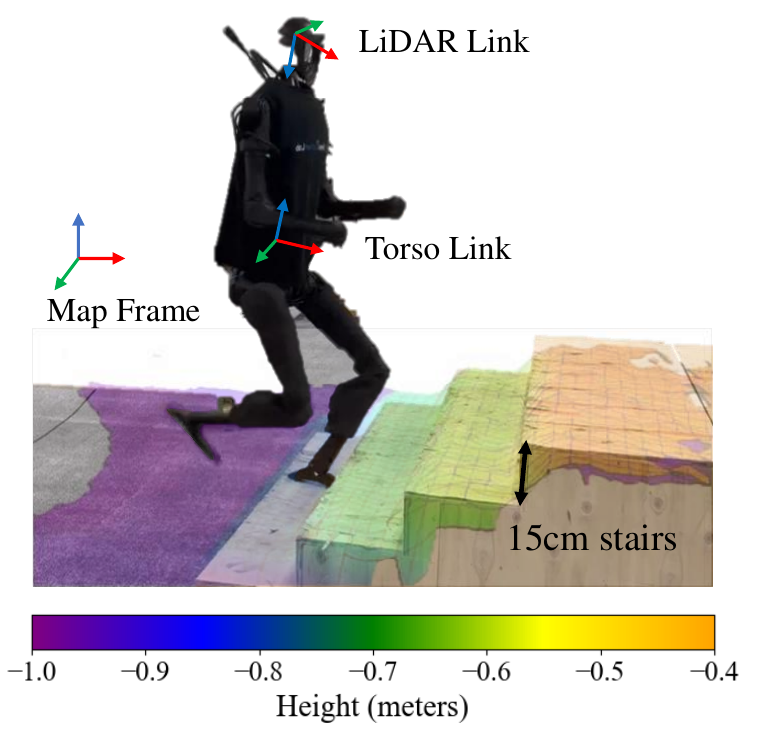}
    \caption{Terrian Perception module implemented by a single LiDAR. The map frame overlaps the initial position of the torso link and is fixed with the world. The colored grid map demonstrates the height of the terrain in the map frame.}
    \label{fig:terrian_perception}
\end{figure}
In the real-world experiments, we employ the elevation map module~\cite{miki2022elevation} to capture perceptive observations as demonstrated in Fig. \ref{fig:terrian_perception}. It requires both odometry and the ground point cloud to generate a grid-based representation of ground heights. Unlike applications involving quadrupeds, it is critical to ensure that the z-axis of the map frame aligns with the direction of gravity. Thus, we implement an initial rotation of the map frame to ensure alignment with the average acceleration direction(i.e. gravity direction). Additionally, the point cloud is processed to filter out points based on height, preserving only the ground points. The filtered point cloud and odometry data are then inputted into the elevation map module to produce the grid-based height map in the map frame. We test elevation maps with two sensor setups: the first setup utilized a single Mid-360 LiDAR to provide both odometry, integrated via FAST-LIO~\cite{xu2021fast,xu2022fast}, and point cloud data. The second setup employed a Realsense T265 camera for odometry and a Realsense D435 camera for point cloud acquisition. While both setups provide accurate elevation maps during stable locomotion, the LiDAR-based setup demonstrated enhanced robustness in scenarios involving rapid or irregular movements.

\paragraph{Elevation Sampling}
We sample 96 points in the frame of the robot aligning its z-axis to the negative direction of gravity. The points are distributed in a $0.8m \times 1.2m$ square with the robot as the center. These points' relative z coordinates to the base link are used as our perceptive input.

\paragraph{Terrain-based Perceptive Internal Model}
To make the most of perceptive information, we not only let the policy use it for foothold planning but also use it for state estimation, which we concluded as PIM (Perceptive Internal Model). PIM is similar to HIM, however, instead of using proprioceptive observation history to estimate the robot's next state, current perceptive observation $\mathbf{o}^{p}_{t}$ is concatenated with the proprioceptive observation history $\mathbf{o}^{n}_{t-H: t}$ to provide more information for more accurate estimation of next-step state since terrain plays an important role in robot state transfer. The effectiveness is shown in Section \ref{sec:pim_vs_hum}.

\subsection{Action Curriculum}
We introduce a curriculum on action space to simplify the training. To be more specific, we choose some of the joints that are less important in the locomotion task, i.e., joints of the robot's arm, and joints of the robot's waist. We set the range of these joints to zero and increase their range as the training goes on.

\subsection{Symmetry Regularization}
We also introduce a symmetry regularization technique similar to \cite{su2024leveraging} to improve the harmony of the gait. In detail, we involve three operators: $G_o^{n}$ which flips the robot's proprioceptive observation with respect to the x-z plane, $G_o^{p}$ which flips the robot's perceptive observation with respect to the x-z plane, and $G_a$ which flips the robot's action with respect to the x-z plane. Then during optimization of PIM, we augment the data with $G_o$. During policy optimization, we involve two additional losses:
\begin{equation}
    \begin{aligned}
        \mathcal{L}_{symmetry}^{policy} &= MSE(G_a(\pi(\mathbf{o}^{n}_t, \mathbf{o}^{p}_t, PIM(\mathbf{o}^{n}_{t-H: t}, \mathbf{o}^{p}_{t})),\\
        &\pi(G_o^n(\mathbf{o}^{n}_t), G_o^p(\mathbf{o}^{p}_t), PIM(G_o^n(\mathbf{o}^{n}_{t-H: t}), G_o^p(\mathbf{o}^{p}_t)))\\
        \mathcal{L}_{symmetry}^{value} &= MSE(V(\mathbf{o}^{n}_t, \mathbf{o}^{p}_t), V(G_o^n(\mathbf{o}^{n}_t), G_o^p(\mathbf{o}^{p}_t)))
    \end{aligned}
\end{equation}

\subsection{Training Process}
At each time, the proprioception observation history $\mathbf{o}^{n}_{t-H: t}$ and perceptive observation $\mathbf{o}^{p}_{t}$ are fed into the perceptive internal model to obtain an estimation of the robot's linear velocity $\hat{v}_{t+1}$ and a latent variable $l_{t}$ predicting the robot's next proprioception information. Then they are fed into the policy network with current observation $\mathbf{o}_{t}$ to obtain action $\mathbf{a}_{t}$. During the policy optimization process, the perceptive internal model is frozen and we only update the policy network and value network using PPO. After policy optimization, we will use the collected trajectories to optimize the perceptive internal model.
\subsection{Reward Functions}
Our reward functions guide the robot to follow certain velocity commands and maintain soft contact with the ground. Besides some reward functions borrowed from quadrupedal locomotion, we designed some new rewards for humanoid locomotion. For example, we wish that the distance of two feet is not too close, then we involve the reward called \textbf{feet lateral distance}, which computes the lateral distance of two feet in the robot's frame and penalizes distance closer than $d_{min}$. We also wish that the feet of the humanoid robot are parallel to the ground, so we use a reward function called \textbf{feet ground parallel} by adding five sample points on each foot of the humanoid robot, indicating the front, middle, hind, left and right, then penalize the variance of distances of these sample points to the ground. Detailed reward functions are listed in Table \ref{reward}.
\begin{table}[h]
    \centering
    \caption{Rewards}
    \begin{tabular}{lll}
    \toprule[1.5pt] Reward & Equation $\left(r_i\right)$ & Weight $\left(w_i\right)$ \\
    \midrule[1.5pt] Lin. velocity tracking & $\exp \left\{- \frac{\|\mathbf{v}_{x y}^{\text {cmd }}-\mathbf{v}_{x y}\|_2^2}{\sigma}  \right\}$ & 1.0 \\
    Ang. velocity tracking & $\exp \left\{- \frac{\left(\omega_{\text {yaw }}^{\text {cmd }}-\omega_{\text {yaw }}\right)^2}{\sigma} \right\}$ & 1.0 \\
    Linear velocity $(z)$ & $v_z^2$ & -0.5 \\
    Angular velocity $(x y)$ & $\|\boldsymbol{\omega}_{x y}\|_2^2$ & -0.025 \\
    Orientation & $\|\mathbf{g}_{x}\|_2^2 + \|\mathbf{g}_{y}\|_2^2$ & -1.25 \\
    Joint accelerations & $\|\ddot{\boldsymbol{\theta}}\|_2^2$ & $-2.5 \times 10^{-7}$ \\
    Joint power & $\frac{|\boldsymbol{\tau} \| \dot{\boldsymbol{\theta}}|^{T}}{\|\mathbf{v}\|_2^2 + 0.2 * \|\boldsymbol{\omega}\|_2^2}$ & $-2.5 \times 10^{-5}$ \\
    Body height w.r.t. feet & $\left(h^{\text {target}}-h\right)^2$ & 0.1 \\
    Feet clearance & $\sum\limits_{feet} \left(p_{z}^{\text {target}}-p_{z}^{i}\right)^2 \cdot v_{xy}^{i}$ & -0.25 \\
    Action rate & $\|\mathbf{a}_t-\mathbf{a}_{t-1}\|_2^2$ & -0.01 \\
    Smoothness & $\|\mathbf{a}_t-2 \mathbf{a}_{t-1}+\mathbf{a}_{t-2}\|_2^2$ & -0.01 \\
    Feet stumble & $\mathbf{1}\left\{\exists i,\left|\mathbf{F}_i^{x y}\right|>3\left|F_i^z\right|\right\}$ & -3.0 \\
    Torques & $\sum\limits_{\text{all joints}} |\frac{\tau_i}{kp_{i}}|_{2}^{2}$ & $-2.5 \times 10^{-6}$ \\
    Joint velocity & $\sum\limits_{\text{all joints}} \dot{\theta}_{i}|_{2}^{2}$ & $-1 \times 10^{-4}$ \\
    Joint tracking error & $\sum\limits_{\text{all joints}}|\theta_{i} - \theta^{target}_{i}|^{2}$ & -0.25 \\
    Arm joint deviation & $\sum\limits_{\text{arm joints}}|\theta_{i} - \theta^{default}_{i}|^{2}$ & -0.1 \\
    Hip joint deviation & $\sum\limits_{\text{hip joints}}|\theta_{i} - \theta^{default}_{i}|^{2}$ & -0.5 \\
    Waist joint deviation & $\sum\limits_{\text{waist joints}}|\theta_{i} - \theta^{default}_{i}|^{2}$ & -0.25 \\
    Joint pos limits & $\sum\limits_{\text{all joints}} \text{out}_{i}$ & -2.0 \\
    Joint vel limits & $\sum\limits_{\text{all joints}} RELU(\hat{\theta}_{i} - \hat{\theta}^{max}_{i})$ & -0.1 \\
    Torque limits & $\sum\limits_{\text{all joints}} RELU(\hat{\tau}_{i} - \hat{\tau}^{max}_{i})$ & -0.1 \\
    No fly & $\mathbf{1}\{\text{only one feet on ground}\}$ & 0.25 \\
    Feet lateral distance & $|y_{\text{left foot}}^{B} - y_{\text{right foot}}^{B}| - d_{min}$ & 2.5 \\
    Feet slip & $\sum\limits_{feet}\left|\mathbf{v}_i^{\text {toot }}\right| * \sim \mathbf{1}_{\text {new contact }}$ & -0.25 \\
    Feet ground parallel & $\sum\limits_{feet}Var(H_i)$ & -2.0 \\
    Feet contact force & $\sum\limits_{feet} RELU(F_{i}^{z} - F_{th})$ & $-2.5 \times 10^{-4}$ \\
    Feet parallel & $Var(D)$ & -2.5 \\
    Contact momentum & $\sum\limits_{feet}|v_{i}^{z} * F_{i}^{z}|$ & $-2.5 \times 10^{-4}$ \\
    \bottomrule[1.5pt]
    \end{tabular}
    \label{reward}
\end{table}

\section{Results} \label{results}

\begin{figure}
    \centering
    \includegraphics[width=0.9\linewidth]{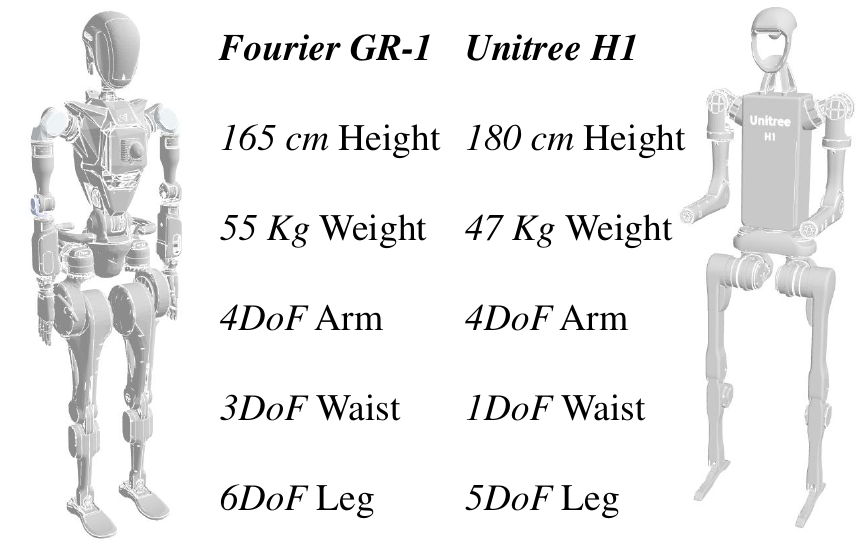}
    \caption{Robot Hardware Setups in terms of Height, Weight, and Degree of Freedom (DoF)}
    \label{fig:robots}
\end{figure}

In the experiments, we validate the efficiency and generalizability of the proposed humanoid locomotion on two humanoid robots: Unitree H1 and Fourier GR-1 (Fig. \ref{fig:robots}), the experiments emphasize:

\begin{itemize}
    \item Effectiveness of Perception in estimating next-step state.
    \item Effectiveness of Perceptive Locomotion: We demonstrate the proposed method successfully navigates extreme terrains including stairs $(\geq 0.15m)$, wooden platforms $(\geq 0.4m)$, slopes $(15 \degree)$, and gaps.
    \item Effectiveness of the Perceptive Internal Model(PIM): The terrain-based internal model accurately predicts the robot's states, enhancing the stability and adaptability of the locomotion.
    \item Cross-Platform Validation: We validate the proposed method on two different humanoid robots in challenging environments, demonstrating its robustness in mastering disturbances and difficult terrains.
\end{itemize}

\subsection{PIM vs HIM}
\label{sec:pim_vs_hum}
\begin{figure}
    \centering
    \includegraphics[width=0.9\linewidth]{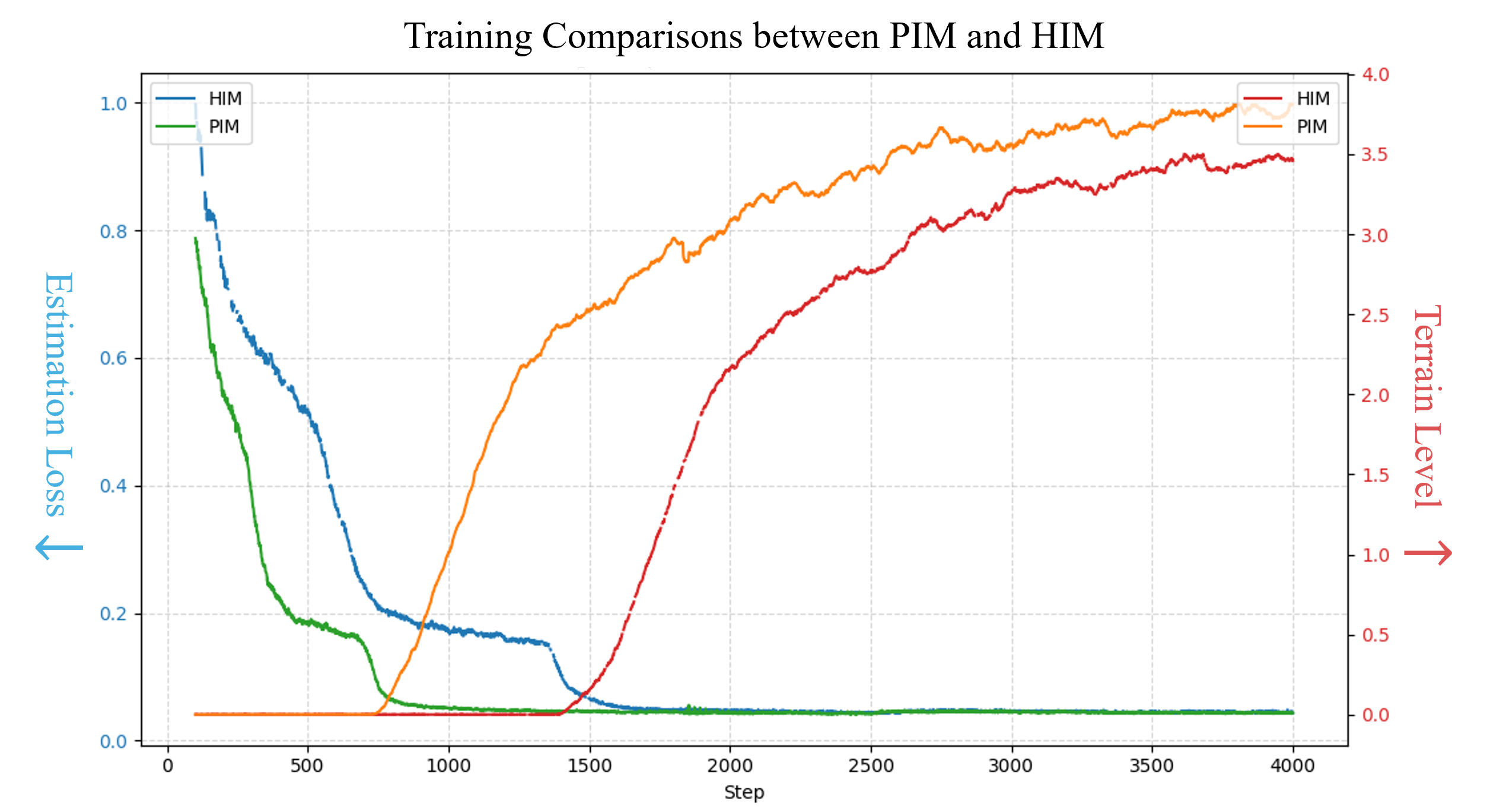}
    \caption{Estimation Loss and Terrain Level During Training}
    \label{fig:HIM vs PIN}
\end{figure}

We compare the estimation accuracy of PIM and HIM, here the difference is that PIM uses perception for estimating while HIM does not, both of their corresponding policies use perception for locomotion. As shown in Fig \ref{fig:HIM vs PIN}, where terrain level indicates the difficulty of terrain that the robot can traverse and estimation loss is the error between estimated velocity and real velocity. PIM can achieve higher training efficiency and provide higher estimation accuracy, as well as enable robots to traverse more difficult terrain. 

\begin{figure*}[!h]
\centering{\includegraphics[width=\textwidth]{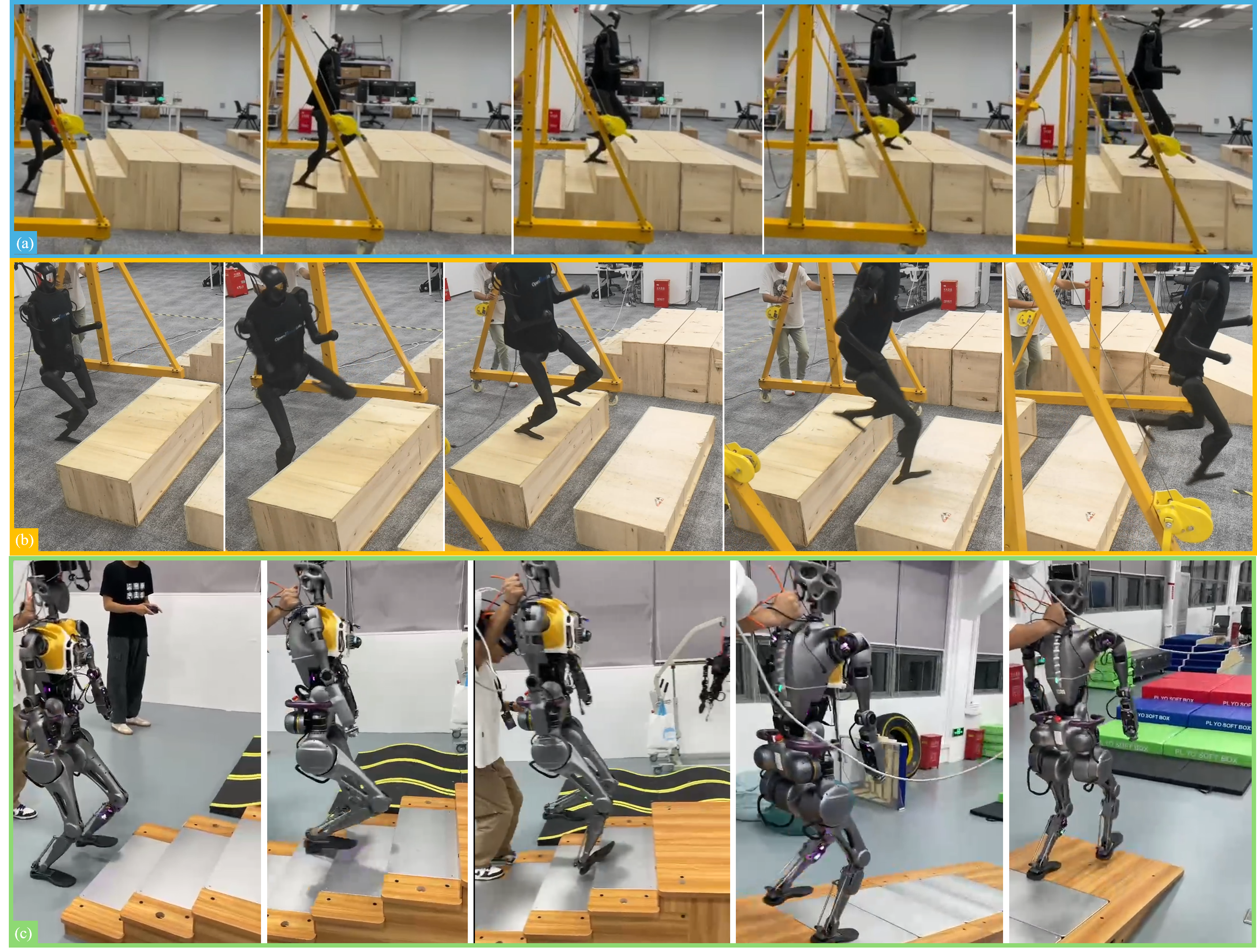}}
\caption{We conducted extensive hardware experiments to validate the effectiveness of the proposed method across different humanoid robots. (a) H1 traverses stairs with heights of 15 cm. (b) H1 consecutively steps onto a wooden platform and jumps over a gap between two platforms. (c) We successfully deploy the proposed policy on GR1 .}
\label{hardware_exp}
\end{figure*}

\subsection{Stairs}
We validate that the proposed method enables the robot to successfully traverse stairs with a height of 15 cm (Fig. \ref{hardware_exp}-a), surpassing existing methods in the literature, such as~\cite{gu2024advancing, liao2024berkeley}, which achieve a maximum height of only 10 cm. This demonstrates the enhanced capability of the proposed perceptive locomotion in handling more challenging terrains. In addition, we would like to highlight the challenge posed by the foot length, which equals the stair width $(30 cm)$. This constraint requires precise foot placement to avoid stepping too far forward, which would cause the foot to get stuck on the next step and prevent proper lifting, or stepping too far back, which would result in a loss of balance and the robot falling. This highlights the critical need for accurate state prediction and perception, further demonstrating the effectiveness of the proposed perceptive locomotion policy in handling challenging environments with precise control.

\subsection{High platform \& gap}
Compared to stair climbing, jumping onto high platforms and between gaps demands less perception accuracy but requires the policy to respond more rapidly to changes in perception. As demonstrated in Fig. \ref{hardware_exp}-b, H1 successfully performs consecutive jumps onto platforms and over gaps, which highlights the effectiveness of the proposed method in perceptive integrated state estimation and executing extreme parkour actions.

\subsection{Cross-Platform Validation}
In addition to H1, we applied the same training process of the proposed method to Fourier GR-1, unlike H1, whose ankle has only one tandem DOF, GR-1 has two DOF at the ankles powered by \textbf{parallel joints}. Our method can still enable it to traverse various terrains as demonstrated in Fig. \ref{hardware_exp}-c. This highlights the method's robustness to different robot configurations, such as variations in height, mass, joint configuration, and foot design. These results suggest the potential of the proposed method to be developed into a generalized solution for humanoid locomotion across diverse platforms.

\subsection{Whole-body Movement}
As a whole-body control method, we observed that the robot performs coordinated upper-body movements to maintain balance without prior knowledge, particularly during extreme actions such as jumping. In general, the robot avoided walking with the same hand or foot during flat-ground tests. During jumping motions, as demonstrated in the fourth image of Fig. \ref{hardware_exp}-b, the robot swings its left arm forward heavily as its right leg jumps, maintaining balance during these extreme actions. This outcome indicates that with appropriate reward design and model prediction as achieved by our method, the robot can generate simple, human-like movements without prior data or imitation learning.

\section{Conclusion} \label{conclusion}
In this work, we propose an effective framework for perceptive humanoid locomotion in complex terrain. The framework includes Perceptive Internal Model(PIM) for next-state prediction, which utilizes perceptive information in addition to HIM. We also present some novel techniques in humanoid locomotion including action curriculum, symmetry regularization, and reward function design. We demonstrate success in sim-to-real deployment on different humanoid robots with different heights, mass, joint configurations, and foot designs. To the best of our knowledge, this work is the first perceptive locomotion framework that works well across different robot platforms.

\bibliographystyle{IEEEtran}

\end{document}